\documentclass[10pt,twocolumn,letterpaper]{article}

\usepackage[pagenumbers]{wacv} % To force page numbers, e.g. for an arXiv version

\newcommand{\ours}{\texttt{ECO-M2F}}

\newcommand{\train}{\mathcal{D}}

\usepackage{graphicx}
\usepackage{color}
\usepackage{placeins}
\usepackage{booktabs}
\usepackage{soul}
\usepackage{nicefrac}
\usepackage{wrapfig, tikz}
\usepackage{amsmath,amsfonts,bm,xspace}
\usepackage{bbm}
\usepackage{color}
\usepackage{enumitem, multirow}
\usepackage{fontawesome5}
\usepackage[many]{tcolorbox}
\usepackage{colortbl}

\newcommand{\inImg}{\vx} % input image
\newcommand{\feature}{\bm{s}} % features from backbone
\newcommand{\gtseg}{\bm{y}} % ground truth segmentation
\newcommand{\idx}{i} % index of input image
\newcommand{\superIdx}{^{(\idx)}} % superscript of image index
\newcommand{\lossCoef}{\alpha} % coefficient of loss
\newcommand \loss {\mathcal{L}} % Loss
\newcommand \NGlayer{K} % # of Pixel Decoder layer 
\newcommand \Glayer{k} % Gating layer
\newcommand \GPRatio{\beta} % ratio of GFlop & PQ
\newcommand \Goutput{\vg} % output of gating function
\newcommand \pqv{\vq} % vector of pq at each layer
\newcommand \tgtIdx{t\superIdx} % gating target of i-th image
\newcommand \uF{u} % utility function
\newcommand \bbF{b} % backbone as a function
\newcommand \pdF{f} % one layer in Pixel decoder as a function
\newcommand \headF{h} % transformer decoder and prediction head as a function
\newcommand \outMF{\tilde{\gtseg}} % M2F

\DeclareMathOperator*{\argmax}{arg\,max}

\newcommand{\stepA}{\textbf{A}}
\newcommand{\stepB}{\textbf{B}}
\newcommand{\stepC}{\textbf{C}}

\def\vg{{\bm{g}}}
\def\vq{{\bm{q}}}
\def\vx{{\bm{x}}}
\def\vz{{\bm{z}}}
\def\mW{{\bm{W}}}
\def\evq{{q}}

\newcommand*\rot{\rotatebox{90}}

\definecolor{wacvblue}{rgb}{0.21,0.49,0.74}
\usepackage[pagebackref,breaklinks,colorlinks,allcolors=wacvblue]{hyperref}

\def\wacvPaperID{41} % *** Enter the WACV Paper ID here
\def\confName{WACV}
\def\confYear{2026}

\title{Image-Specific Adaptation of Transformer Encoders for Compute-Efficient Segmentation}

\author{\textbf{Manyi Yao}$^\dagger$, \textbf{Abhishek Aich}$^\ddagger$, \textbf{Yumin Suh}$^\ddagger$, \textbf{Amit Roy-Chowdhury}$^\dagger$,\\ \textbf{Christian Shelton}$^\dagger$, \textbf{Manmohan Chandraker}$^{\ddagger,\star}$\\
$^\ddagger$NEC Laboratories, America, $^\dagger$University of California, Riverside,\\ $^\star$University of California, San Diego}

\begin{document}
\maketitle
\begin{abstract}
Vision transformer-based models bring significant improvements to image segmentation tasks. Although these architectures offer powerful capabilities irrespective of specific segmentation tasks, their use of computational resources can be taxing on deployed devices. 
    % \crscomment{I would replace ``can be significantly optimized'' with ``can be burdensome.'' }
    One way to overcome this challenge is by adapting the computation level to the specific needs of the input image rather than the current one-size-fits-all approach. To this end, we introduce \ours{} or \textbf{E}ffi\textbf{C}ient Transf\textbf{O}rmer Encoders for \textbf{M}ask\textbf{2F}ormer-style models. Noting that the encoder module of M2F-style models incur high resource-intensive computations, \ours{} provides a strategy to self-select the number of hidden layers in the encoder, conditioned on the input image. To enable this self-selection ability for providing a balance between performance and computational efficiency, we present a three-step recipe. The \textit{first} step is to train the parent architecture to enable early exiting from the encoder. The \textit{second} step is to create a derived dataset of the ideal number of encoder layers required for each training example. The \textit{third} step is to use the aforementioned derived dataset to train a gating network that predicts the number of encoder layers to be used, conditioned on input images. Additionally, to change the computational-accuracy trade-off, only steps two and three need to be repeated which significantly reduces retraining time. Experiments on the public datasets show that the proposed approach reduces expected encoder computational cost while maintaining performance, adapts to various user compute resources, is flexible in architecture configurations, and can be extended beyond the segmentation task to object detection. 
\end{abstract}

\section{Introduction}
\label{sec:intro}
With the advent of powerful \textit{universal} image segmentation architectures \cite{cheng2021per, cheng2021mask2former, jain2023oneformer, gu2024dataseg}, it is highly desirable to prioritize the computational efficiency of these architectures for their enhanced scalability, \eg, use on resource-limited edge devices. These architectures are extremely useful in tackling instance \cite{he2017mask}, semantic \cite{tu2008auto}, and panoptic \cite{kirillov2019panoptic} segmentation tasks using one generalized architecture, owing to the transformer-based \cite{vaswani2017attention} modules. These universal architectures leverage DEtection TRansformers or DETR-style \cite{carion2020end} modules and represent both \textit{stuff} and \textit{things} categories \cite{kirillov2019panoptic} using general feature tokens. This is an incredible advantage over preceding segmentation methods \cite{sun2023remax, hu2023you, xu2024rap} in the literature that require careful considerations in design specifications. Hence, these segmentation architectures reduce the need for task-specific choices that favor the performance of one task over the other \cite{cheng2021mask2former}.
\begin{figure}[t]
    \centering
    \includegraphics[width=0.95\columnwidth]{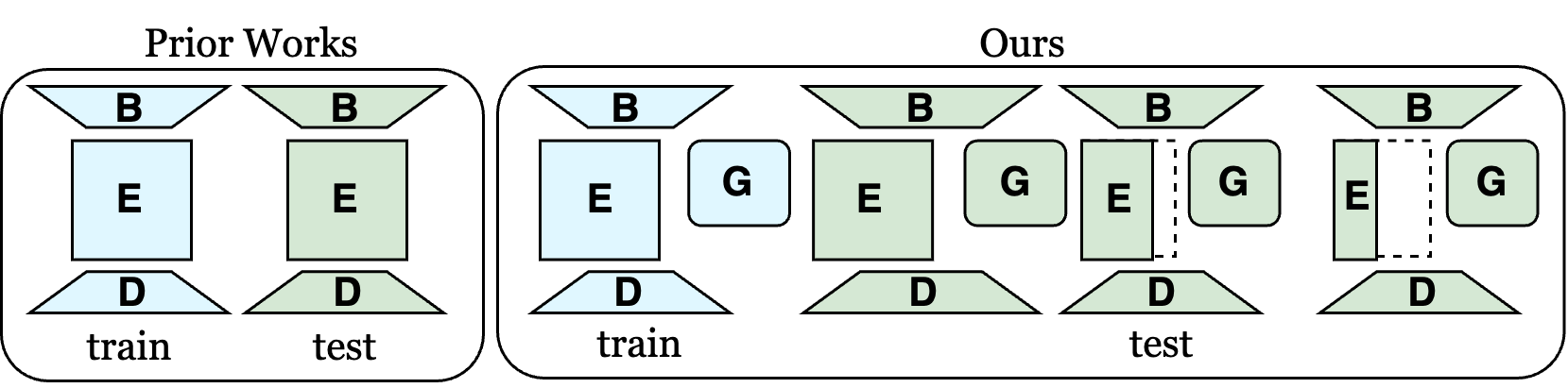}
    \caption{\textbf{Comparison to prior works.} Instead of conventional M2F-style architecture that provides a ``one-size-fits-all'' solution, our method {\ours} trains models to run directlys at various resource encoder depths by leveraging a gating function. \textbf{B}, \textbf{E}, \textbf{D}, and \textbf{G} denote the backbone, encoder, decoder, and (our proposed) gating network, respectively.} %
      \label{fig:teaser_a}
\end{figure}

\begin{figure}[t]
  \centering
  \begin{minipage}[b]{0.50\columnwidth}
      \centering
      \includegraphics[width=\textwidth]{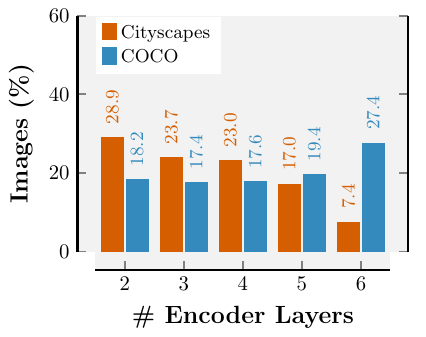}
      (a) % \caption*{(a)}
      \label{fig:teaser_b}
  \end{minipage}
  \begin{minipage}[b]{0.48\columnwidth}
      \centering
      \includegraphics[width=\textwidth]{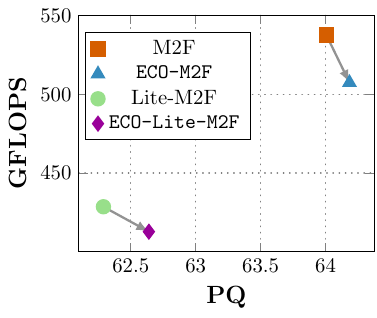}
      (b) % \caption*{(b)}
      \label{fig:teaser_CS}
  \end{minipage}
  \caption{(a) Histogram of images achieving best panoptic segmentation by the number of encoder layers. (b) Our method demonstrates superior performance and lower computational cost compared to the baseline models. (Dataset: Cityscapes; Backbone: SWIN-T)}

  \label{fig:teaser_combined}
\end{figure}

% What are the existing methods and where do they lack?
State-of-the-art models for universal segmentation like Mask2former (M2F) \cite{cheng2021mask2former} are built on the key idea inspired by DETR: ``mask'' classification is versatile enough to address both semantic- and instance-level segmentation tasks. However, the problem of efficient M2F-style architectures has been underexplored. With backbone architectures (\eg, Resnet-50 \cite{he2016deep}, SWIN-Tiny \cite{liu2021swin}), \cite{li2023lite} showed that DETR-style models incur the highest computations from the transformer encoder due to maintaining full-length token representations from multi-scale backbone features. While existing works like \cite{li2023lite, lv2023detrs} primarily focus on scaling the input token to improve efficiency, this approach often neglects other aspects of model optimization and leads to a ``one-size-fits-all'' solution (\cref{fig:teaser_a}). This limitation leaves significant room for further efficiency improvements.

% What is our method and what are the components?
Given this growing importance of M2F-style architectures and the indispensable need for efficiency for real-world deployment, we introduce \ours{} or `\textbf{E}ffi\textbf{C}ient Transf\textbf{O}rmer Encoders' for M2F-style architectures. Our key idea comes from our observation made on the training set of COCO \cite{lin2014microsoft} and Cityscapes \cite{cordts2016cityscapes} dataset demonstrated in \cref{fig:teaser_combined}~(a). We plot a histogram of the number of hidden encoder layers that produces the best panoptic segmentation quality \cite{kirillov2019panoptic} for each image. It can be seen that not all images require the use of all $\NGlayer$ hidden layers of the transformer encoder in order to achieve the maximum panoptic segmentation quality \cite{kirillov2019panoptic}. With this insight, we propose to create a dynamic transformer encoder that economically uses the hidden layers, guided by a gating network that can select different depths for different images.

To achieve the aforementioned ability, \ours{} leverages the well-studied early exiting strategy \cite{tang2023you, xu2023lgvit, liu2021mevt, wang2022single, jiang2023multi, yang2023exploiting, valade2024eero, tang2023need, zhang2023adaptive} to create stochastic depths for the transformer encoder to improve inference efficiency. Previous exit mechanisms have primarily relied on confidence scores or uncertainty scores, typically applied in classification tasks. However, implementing such mechanisms in our context would necessitate the inclusion of a decoder and a prediction head to generate a reliable confidence score. This additional complexity introduces a significant number of FLOPs, rendering it impractical for our purposes. By contrast, \ours{} provides a three-step training recipe that can be used to customize the transformer encoder on the fly given the input image. Step~\stepA{} involves training the parent model to \textit{be dynamic} by allowing stochastic depths at the transformer encoder. Using the fact that the transformer encoder maintains the token length of the input throughout the hidden layers  constant,  Step~\stepB{} involves creating a \textit{Derived} dataset from the training dataset whose each sample contains a pair of images and layer number that provides the highest segmentation quality. Finally, Step~\stepC{} involves training a \textit{Gating Network} using the derived dataset, whose function is to decide the number of layers to be used given the input image. 

The key contributions of {\ours} are manifold:

\begin{itemize}
    % $\bullet$ 
    \item \textit{Enhanced Encoder Flexibility}: It fine-tunes pre-trained M2F-style architectures for early exits from the encoder, leveraging the constant token length in hidden layers.
    % \\$\bullet$ 
    \item \textit{Adaptive Gating Network Training}: It introduces a Gating network to optimize encoder layer usage, allowing the architecture to determine the optimal layer count per input without performance loss or confidence thresholds.
    % \\$\bullet$ 
    \item \textit{Innovative Efficient Training}: The Gating network's training approach enables the architecture to adapt to varying computational budgets, with the cost of only Step~\stepC (about 2.5\% of Step~\stepA on COCO \cite{lin2014microsoft} dataset).
    % \\$\bullet$ 
    \item \textit{Integration and Expansion}: {\ours} can incorporate recent advancements in token length scaling for transformer encoder efficiency and extend its application open-vocabulary segmentation and detection tasks.
    % \\$\bullet$ 
    \item \textit{Competitive Performance-Computational Trade-off}: Our method strikes a competitive balance between performance and computational efficiency, and can even excel in both areas simultaneously, as illustrated in \cref{fig:teaser_combined}~(b).

\end{itemize}

\section{Related Works}
\label{sec:related_works}
%
% \amit{It may be better to have a para on the complexity of transformer architectures in general, then one specific to segmentation. The first few sentences of the first para are written from a more general perspective. This goes back to my earlier point about how we frame the contribution of the paper.} \abhi{Discuss during meeting.}
\paragraph{Efficient image segmentation.} With the rise of transformers \cite{vaswani2017attention}, researchers are increasingly interested in creating image segmentation models that work effectively in various settings, without requiring segmentation type-specific modifications to the model itself. Building on DETR \cite{carion2020end}, multiple universal segmentation architectures were proposed \cite{cheng2021per,cheng2021mask2former, jain2023oneformer, gu2024dataseg} that use a transformer decoder to predict masks for each entity in the input image. However, despite the significant progress in overall performance across various tasks, these models still face challenges in deployment on resource-constrained devices. Current emphasis \cite{cheng2020panoptic, fan2021rethinking, hou2020real, hu2023you, 10296714, xu2023pidnet, yu2018bisenet, yu2021bisenet} for efficiency for image segmentation has mostly been on specialized architectures tailored to a single segmentation task. Unlike these preceding works, \ours makes no such assumption on the segmentation task and addresses the limitation of inefficiency in M2F-style universal architectures that are task-agnostic.
\paragraph{Early-exiting in vision transformers.} %Recent years have witnessed a surge in research aimed at improving the inference efficiency of large-scale transformer models \cite{wan2023efficient, xu2024survey, tang2023you, xu2023lgvit, liu2021mevt, wang2022single, jiang2023multi, yang2023exploiting, valade2024eero, tang2023need, zhang2023adaptive} where early exiting strategies have been much explored.
Recent works on early exiting \cite{wan2023efficient, xu2024survey, tang2023you, xu2023lgvit, liu2021mevt, wang2022single, jiang2023multi, yang2023exploiting, valade2024eero, tang2023need, zhang2023adaptive} aim to boost inference efficiency for large transformers. Some works \cite{xu2023lgvit, liu2021mevt, tang2023you} used early exiting for classification tasks along with manually chosen confidence threshold in vision transformers. For example, \cite{xu2023lgvit} proposed an early exiting framework for classification task ViTs combining heterogeneous task heads. Similarly, \cite{tang2023you} proposed an early exiting strategy for vision-language models by measuring layer-wise similarities by checking multiple times to exit early.  Applying early exiting solely to the encoder (like \cite{xu2023lgvit}) is infeasible due to the dependency on separate decoders, leading to an unacceptable optimization load. In contrast, methods like \cite{tang2023you} suffer from redundant computations for exit decisions at all possible choices, hindering efficient resource allocation. In contrast, \ours only trains one decoder for all possible exit routes, as well as uses a gating module to decide the number of encoder layers required for the model, depending on the input image. 
\section{Proposed Methodology: \ours}
\label{sec:method}
\subsection{Model Preliminaries}
% \paragraph{Model preliminaries.} 
\noindent We first review the meta-architecture of M2F \cite{cheng2021mask2former} upon which {\ours} is based, along with the notation. This class of models contains:\\
\begin{itemize}
    % $\bullet$ 
    \item a \emph{backbone} $\bbF(\cdot)$ which takes the $\idx$-th image $\inImg\superIdx$ as input to generate multi-scale feature maps $\bbF(\inImg\superIdx)$, represented as ${\feature_1, \feature_2, \feature_3, \feature_4}$. These multi-scale feature maps correspond to spatial resolutions typically set at $\nicefrac{1}{32}$, $\nicefrac{1}{16}$, $\nicefrac{1}{8}$, and $\nicefrac{1}{4}$ of the original image size, respectively.    
    % \\$\bullet$ 
    \item a \emph{transformer encoder} (called the ``pixel decoder'' \cite{cheng2021mask2former}), which is composed of multiple layers of transformer encoders. The function of this module is to generate rich token representation from $\{\feature_1, \feature_2, \feature_3\}$ and generate per-pixel embeddings from $\feature_4$. Each layer in the transformer encoder, denoted as $\pdF_\Glayer(\cdot)$ (where $\Glayer\in\{1,2,\dots,\NGlayer\}$) is successively applied to $\bbF(\inImg\superIdx)$, with $\pdF_\NGlayer(\cdot)$ being the last layer in the transformer encoder.
    % \\$\bullet$ 
    \item a \emph{transformer decoder} (along with a segmentation head) that takes two inputs: the output of the transformer encoder and the object queries. The object queries are decoded to output a binary mask along with the corresponding class label. 
\end{itemize}

For brevity, we collectively refer to the operations in the transformer decoder and segmentation head together as $\headF(\cdot)$. Thus, the output of the meta-architecture with $K$ encoder layers (a predicted mask $\outMF_\NGlayer\superIdx$ and corresponding label $\tilde{\ell}_\NGlayer\superIdx$) can be written as 
\begin{equation}
    \{\outMF_\NGlayer\superIdx, \tilde{\ell}_\NGlayer\superIdx\} = h\circ f_\NGlayer\circ\cdots\circ f_2\circ f_1\circ b(\inImg\superIdx)\,.
\end{equation}

Here, the operation $\circ$ represents function composition, \eg, $g\circ f(x) = g(f(x))$, and subscript denotes output predicted using $K$ encoder layers. With $\{\gtseg\superIdx, \ell\superIdx\}$ as the pair of ground truth segmentation map and corresponding label of image $\inImg\superIdx$, the final loss \cite{cheng2021mask2former} is computed as
\begin{align}
    \loss_\NGlayer = \lambda_{\text{mask}}\loss_{\text{mask}}(\outMF_\NGlayer\superIdx, \gtseg\superIdx) + \lambda_{\text{class}}\loss_{\text{class}}(\tilde{\ell}_\NGlayer\superIdx, \ell\superIdx )\,,
    \label{eq:orig_loss}
\end{align}
where $\loss_{\text{mask}}(\cdot,\cdot)$ is a binary mask loss and $\loss_{\text{class}}(\cdot,\cdot)$ is the corresponding classification loss. $\lambda_{\text{mask}}$ and $\lambda_{\text{class}}$ represent the associated loss weights.

\subsection{Method Motivation}
% \paragraph{Method motivation.} 
\noindent Our motivation stems from the observation that layers within the transformer encoder of M2F exhibit non-uniform contributions to Panoptic Quality (PQ) \cite{kirillov2019panoptic}, as discussed in \cref{sec:intro}.
% As illustrated in \cref{fig:PQ_vs_layers}, our analysis on COCO and Cityscapes datasets demonstrates the PQ trends across models featuring from 1 to the original 6-layer configuration. While an increase in layers generally correlates with higher PQ, notable PQ levels are retained even in models with as few as 2 layers. 
This prompts us to question the necessity of all $K=6$ layers for every image and target minimizing layer usage according to the user's computational constraints while ensuring that overall performance remains within acceptable bounds. Hence, we adopt an adaptive early exiting approach driven by three critical components:

1. \emph{Model suitability for early exiting.} Traditional early exiting techniques \cite{tang2023you, xu2023lgvit, liu2021mevt, wang2022single, jiang2023multi, yang2023exploiting, valade2024eero, tang2023need, zhang2023adaptive} often face challenges in maintaining satisfactory performance levels at potential exit points throughout the neural network. We recognize the importance of a model architecture that not only allows for early exiting but also ensures that the performance remains consistently high. Therefore, we aim to develop a model that not only permits early exits but also for which the accuracy steadily improves as the network delves deeper into its architecture. By prioritizing this aspect, we seek to establish a framework where early exiting does not compromise the overall performance of the model.

2. \emph{Efficient and effective gating network for optimal exit decision-making.} The efficacy of an early exiting strategy heavily depends on the ability to make informed exit decisions. A gating network must strike a delicate balance, minimizing computational overhead while effectively identifying components that can be bypassed without compromising accuracy. Our objective is to design a lightweight yet powerful gating mechanism capable of discerning optimal exit points within the model architecture.

3. \emph{Dynamic control mechanism for cost-performance trade-off.} We require a mechanism with the ability to adaptively regulate the balance between computational cost and performance according to user-defined priorities. Such a mechanism empowers the model to exit at the optimal layer based on specific needs and desired outcomes, ensuring efficient resource allocation and maximizing utility in various application scenarios, particularly in resource-constrained environments like edge computing or real-time applications.

Driven by these considerations, {\ours} offers a novel training process that enables an adaptive early exiting mechanism designed to bolster computational efficiency while preserving satisfactory model accuracy. For better understanding, we'll begin with a general overview of model training and inference before diving into the specific details of our training process. 

\paragraph{Training and Inference Overview}
% \paragraph{Training and Inference overview.}

\begin{figure*}[!t]
  \centering
  \includegraphics[trim={0 0 1.5em 0},width=\textwidth]{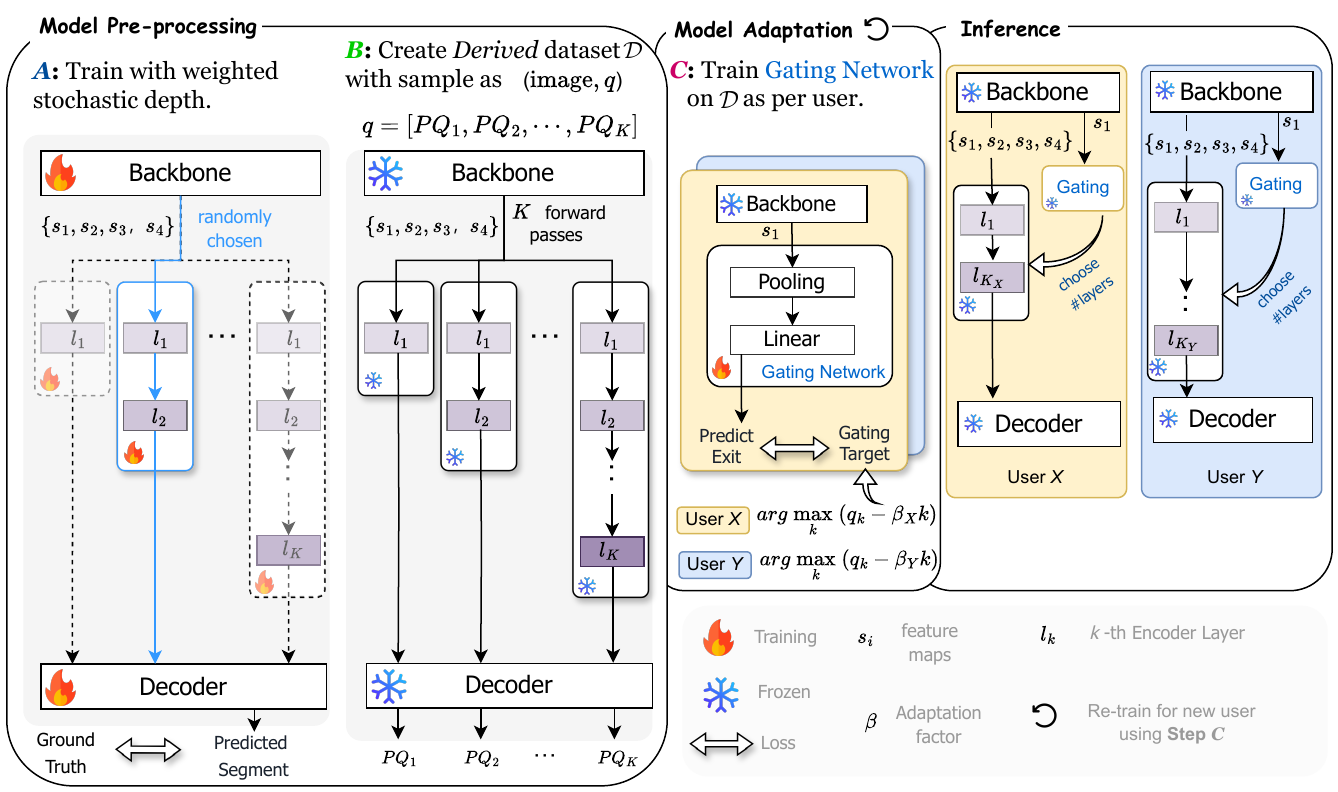}
  \caption{\textbf{{\ours} framework.} During the \textit{model pre-processing} phase, we train the model to exit stochastically at $\NGlayer$ potential exits using Step~\stepA.  Next, in Step~\stepB, we use this model to perform inference on the training images at each exit to create a dataset $\train$. In the \textit{model adaptation} phase, we perform {Step~\stepC} to establish a gating target based on the computational budget and train a lightweight gating network. During \textit{inference}, the network exists at the layer designated by the gating network.}
  \label{fig:main_framework} 
\end{figure*}

\noindent As shown in \cref{fig:main_framework}, the training phase of {\ours} comprises three main steps: 
\begin{enumerate}
    \item Step~\stepA: Train the parent model for early exit via the transformer encoder.
    \item Step~\stepB: Derive a dataset (which we call the Derived dataset) from the dynamic model obtained in Step~\stepA.
    \item Step~\stepC: Train the \textit{Gating Network} to learn optimal exit points in the encoder tailored to users' requirements.
\end{enumerate}

We refer to Step~\stepA and {\stepB} together as \textit{model pre-processing} and Step~\stepC as \textit{model adaptation}. The former is required only once, whereas the latter is repeated as per user requirements. All these steps use the training data subset. 

During inference, the gating network guides the parent model by selecting the optimal exit point based on features extracted from the backbone with just one forward pass for final predictions.

\subsection{Training}
\label{sec:training}
\paragraph{Step~\stepA: Training with Weighted Stochastic Depth}
\label{sec:SDwithP}
In this step, we enable the model to allow exiting at the encoder. To maintain consistently high performance at each exit point, we input each stochastic depth's output to a shared transformer decoder. We then apply \cref{eq:orig_loss} to compute the loss $\loss_\Glayer$ for each exit point $\Glayer$. However, we observe that direct training in this fashion does not encourage the model to use fewer layers to extract and prioritize informative representations, as shown in the experiment results. To address this, we introduce a set of coefficients $\lossCoef_\Glayer$ to emphasize the quality of representations at later layers more, enabling earlier layers to also concentrate on producing effective intermediate representations. As the layer depth increases, the corresponding coefficient $\lossCoef_\Glayer$ grows, ensuring a progressively stricter standard for feature quality. The new loss function is then expressed as
\begin{equation}
    \label{eq:loss}
    \loss_{\text{total}} = \frac{1}{N}\sum_\idx^N\sum_{\Glayer}^\NGlayer \lossCoef_\Glayer\loss_\Glayer,\,\,{\text{where}}\ \forall \Glayer<\Glayer', \lossCoef_\Glayer<\lossCoef_{\Glayer'}\,,
\end{equation}
where $N$ is the number of images in the training set, and $\loss_\Glayer$ is from \cref{eq:orig_loss}.

\paragraph{Step~\stepB: Deriving the Gating Network Training Set}
To facilitate informed exit decisions during inference, our approach is to train a gating network to learn optimal exit strategies. In this step, we facilitate this gating network training by first deriving an intermediate dataset. 

To this end, we record the performance of the pre-trained stochastic depth model (obtained from Step~\stepA) at all potential exit points for each image within the training dataset and create a \textit{Derived} dataset $\train$. Specifically, we associate the $\idx$-th input image $\inImg\superIdx$ with a vector $\pqv\superIdx$ of length $\NGlayer$. Each element $\evq_\Glayer\superIdx$ of $\pqv\superIdx$ represents the predicted panoptic quality \cite{kirillov2019panoptic} upon exiting at the encoder layer $\Glayer$. Hence, each sample of $\train$ can be represented as $ (\inImg\superIdx, \pqv\superIdx)$.

\begin{figure}[t]
    \centering
    \includegraphics[width=0.75\columnwidth]{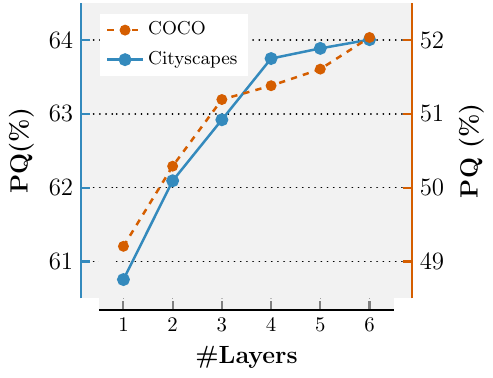} 
    \caption{\textbf{Intuition for \cref{eq:utility_func}}. This figure shows that prioritizing PQ requires more encoder layers, while fewer layers lead to poorer PQ. (Backbone: SWIN-T; training set).}
    \label{fig:PQ_vs_layers}
\end{figure}

\paragraph{Step~\stepC: Training for Gating Network}
\label{sec:gating}
In this step, we train the gating network on dataset $\train$ (obtained from Step~\stepB) to self-select the number of encoder layers based on the input image. Ideally, this module should allow exiting at the encoder layer, which would result in the highest quality segmentation map. With this in mind, we first establish the target exit for the gating network. Note that the panoptic quality generally increases with increasing encoder layers (see \cref{fig:PQ_vs_layers}). However, we would like the gating network to prioritize increasing the panoptic quality while also reducing the number of layers (to reduce the overall computations).  
Consequently, we introduce a utility function expressed as a linear combination of segmentation quality and the depth of the network. This function is formulated as
\begin{equation}
    \uF(\Glayer) = q_\Glayer\superIdx - \GPRatio\Glayer\,,
    \label{eq:utility_func}
\end{equation}
where $\GPRatio$ serves as an \textit{adaptation factor} governing the trade-off between segmentation quality and computational cost. Clearly, a higher value of $\GPRatio$ signifies a greater emphasis on efficiency over segmentation quality. Using \cref{eq:utility_func}, we determine a target exit point $\tgtIdx$ for each image $\inImg\superIdx$ using 
\begin{equation}
    \label{eq:target}
    \tgtIdx = \argmax_\Glayer (\uF(\Glayer))\,.
\end{equation}
With a target designated for each image using \cref{eq:target}, the gating decision can be approached as a straightforward classification problem.
The gating architecture consists of a pooling operation $\vz(\cdot)$ on the token length dimension followed by a linear layer with weights $\mW$. Its output logits can be represented as 
\begin{align}
    \label{eq:gating}
    \Goutput\superIdx &= \mW \vz(\feature_1\superIdx)\,.
\end{align}

In consideration of having minimal impact on the computations due to the gating network, we use the output of the lowest resolution feature map $\feature_1$ as input to the pooling operation. To optimize the gating network, we use the standard cross-entropy loss between the output logits $\Goutput\superIdx$ and the one-hot version of target exit $\tgtIdx$ as our training objective. During inference, the gating network identifies the layer with the highest predicted logits, \ie, $\argmax_\Glayer(g_\Glayer\superIdx)$, as the optimal exit layer for image $\inImg\superIdx$. Note that while there can be more complex choices for the gating network, our simple linear layer in \cref{eq:gating} works well in experiments.

\paragraph{Saving Training Costs through Step~\stepC}
 {\ours} presents a distinct advantage in terms of its adaptability to varying computational constraints. In scenarios where a smaller model is desired, {\ours} necessitates training solely the gating network (\ie, repeat Step~\stepC). Assuming that the computational load is proportional to the depth of the network, \cref{eq:utility_func} enables us to weigh the performance gain against the computational overhead for each exit layer. We achieve this by setting the total number of layers $\NGlayer$ to a smaller number depending on user preferences. For instance, as illustrated in \cref{fig:main_framework}, User $X$ preferring a smaller model compared to User $Y$ may opt for a smaller $\NGlayer$, \ie, $\NGlayer_X<\NGlayer_Y$. Then, given the importance of segmentation quality, we choose $\beta$. With these two variables set in \cref{eq:target}, we train the gating network. This capability shows that {\ours} is versatile and resource-efficient as it adapts to diverse needs and optimizes allocations.

\subsection{Inference}
\noindent In the inference phase, the gating network guides the parent model toward an optimal exit point tailored to each input image. Similar to the training phase, the gating mechanism receives low-resolution features from the backbone and produces a vector of length $\NGlayer$ for each image. The value of $\NGlayer$ remains consistent with that determined in Step~\stepC. Subsequently, the gating network identifies the layer with the highest predicted logits as the optimal exit layer for each image. The parent model adheres to this decision, exiting at the determined layer, and subsequently progresses through the subsequent components to make the final prediction. This dynamic process ensures that the model adaptively selects the most optimal layer for exit during inference, enhancing its efficiency in handling diverse input data.

\label{sec:exp}
\subsection{Experiments Settings}
\paragraph{Datasets} Our study illustrates the adaptability of {\ours} in dynamically managing the trade-off between computation and performance based on M2F \cite{cheng2021mask2former} meta-architecture. We do this on two widely used image segmentation datasets: COCO \cite{lin2014microsoft} and Cityscapes \cite{cordts2016cityscapes}. COCO comprises 80 ``things'' and 53 ``stuff'' categories, with 118k training images and 5k validation images. Cityscapes consists of 8 ``things'' and 11 ``stuff'' categories, with approximately 3k training images and 500 validation images. The evaluation is conducted over the union of ``things'' and ``stuff'' categories.

\paragraph{Evaluation Metrics} We follow the evaluation setting of \cite{cheng2021mask2former} for evaluation of ``universal'' segmentation, \ie, we train the model solely with panoptic segmentation annotations but evaluate it for panoptic, semantic, and instance segmentation tasks. We use the standard \textbf{PQ} (Panoptic Quality \cite{kirillov2019panoptic}) metric to evaluate panoptic segmentation performance. We report \textbf{AP}$_p$ (Average Precision \cite{lin2014microsoft}) computed across all categories for instance segmentation, and \textbf{mIOU}$_p$(mean Intersection over Union \cite{everingham2015pascal}) for semantic segmentation by merging instance masks from the same category. The subscript $p$ denotes that these metrics are computed for the model trained solely with panoptic segmentation annotations. In terms of computational cost, we use GFLOPs calculated as the average GFLOPs across all validation images. All models are trained on the \textit{train} split and evaluated on the \textit{validation} split.

\paragraph{Baseline Models}
We compare \ours{} with two sets of efficient segmentation methods. \textit{First}, we compare with our baseline  universal segmentation architecture M2F \cite{cheng2021mask2former}. Further, we also integrate recently proposed transformer encoder designs (Lite-DETR \cite{li2023lite} and RT-DETR \cite{lv2023detrs}) for efficient object detection into M2F and named them Lite-M2F and RT-M2F, respectively. \textit{Second}, we include comparisons with recent efficient architectures that proposed task-specific components, namely YOSO \cite{hu2023you}, RAP-SAM \cite{xu2024rap}, and ReMax \cite{sun2023remax}.

\paragraph{Architecture Details}
 We focus on standard backbones Res50 \cite{he2016deep} and SWIN-Tiny \cite{liu2021swin} pre-trained on ImageNet-1K \cite{deng2009imagenet}, unless specified otherwise. We set the total number of encoder layers to be 6 following \cite{cheng2021mask2former}. We consider layers 2 to 6 as potential exit points, unless stated otherwise. In our gating network, we use a straightforward 1D adaptive average pooling operation as our pooling function. 

\paragraph{Training Settings}
The experimental setup closely mirrors that of M2F \cite{cheng2021mask2former}, with all model configurations and training specifics following identical procedures. 
We use Detectron2 \cite{wu2019detectron2} and PyTorch\cite{paszke2019pytorch} for our implementation.
For the stochastic depth training phase (Step~\stepA), we initialize weights as provided by M2F and subsequently train 50 epochs for the COCO dataset and 90k iterations for Cityscapes, with a batch size of 16.
For the training of the gating network (Step~\stepC), we perform 2 epochs of training on the COCO dataset and 20k iterations on the Cityscapes dataset, employing the Adam optimizer \cite{kingma2017adam}. The adaptation factor $\GPRatio$ in the utility function, as discussed in \cref{sec:method}, is set to 0.0005 for COCO and 0.003 for Cityscapes, unless otherwise specified. Distributed training is performed using 8 A6000 GPUs. On the COCO dataset, the training time of Step~\stepA{} is 280 GPU hours, Step~\stepB{} is 17 GPU hours, and Step~\stepC{} 7.2 GPU hours. Similarly for Cityscapes dataset, the training time of Step~\stepA{} is 45 GPU hours, Step~\stepB{} is 1 GPU hours, and Step~\stepC{} is 7.2 GPU hours. In Step~\stepA, we use identical settings as M2F for the loss between the predicted segment and ground truth segment, \ie, $\loss_\Glayer$. The weight $\lambda_{\text{mask}}$ is fixed at 5.0, while $\lambda_{\text{class}}$ is set to 2.0 for all classes, except 0.1 for the ``no object'' class. 

\subsection{Main Results}
\begin{table}[!t]
    \begin{center}
        \resizebox{\columnwidth}{!}{
        \begin{tabular}{lcccccccc}
        \toprule
         & \multicolumn{3}{c}{\textbf{Performance} ($\uparrow$)} && \multicolumn{2}{c}{\textbf{GFLOPs} ($\downarrow$)} \\ \cline{2-4} \cline{6-7}
        \multirow{-2}{*}{\textbf{Model}} & \textbf{PQ} & \textbf{mIoU}$_p$ & \textbf{AP}$_p$ && \textbf{Total} & \textbf{Encoder} \\ 
        \midrule
        
        \multicolumn{7}{c}{\textbf{Backbone}: SWIN-T}\\
        \hline
        RT-M2F \cite{lv2023detrs} & 41.36 & 61.54 & 24.68 && 158.30 & 59.66 \\
        Lite-M2F \cite{li2023lite} & 52.70 & 63.08 & 41.10 && 188.00 & 79.78 \\
        M2F \cite{cheng2021mask2former}  & 52.03 & 62.49 & 42.18 && 235.57  & 121.69 \\
        {\ours}($\GPRatio=0.0005$) & 52.06 & 62.76 & 41.51 && 202.39 & 88.47 \\
        {\ours}($\GPRatio=0.02$) & 50.79 & 62.25 & 39.71 && 181.64 & 67.71 \\
        Lite-{\ours} & 52.84 & 63.23 & 42.18 && 178.43 & 64.42 \\
        \hline
        \multicolumn{7}{c}{\textbf{Backbone}: Res50}\\
        \hline
        %
        % TODO: confirming "anals/FlopCCR50M2F_120977_1gpu.log"
        M2F \cite{cheng2021mask2former} & 51.73 & 61.94  & 41.72 && 229.10 & 135.00 \\
        MF \cite{cheng2021per} & 46.50 & 57.80 & 33.00  && 181.00 & --  \\
        
        PEM \cite{cavagnero2024pem} & 46.38 & 55.95 & 34.25 && 110.90 & -- \\
        YOSO$^\dagger$ \cite{hu2023you} & 48.40 & 58.74 & 36.87 && 114.50 & --  \\
        RAP-SAM$^\dagger$\cite{xu2024rap} & 46.90 & -- & -- && 123.00 & --  \\
    
        ReMax$^\dagger$ \cite{sun2023remax} & 53.50 & -- & -- && --  & --  \\
        {\ours} & 51.89 & 61.07 & 41.25 && 195.55 & 92.37 \\
        
        \bottomrule
        \end{tabular}}
    \end{center}
    \caption{\textbf{Evaluation on COCO dataset.\label{tab:result_coco}}. Our method achieves competitive performance with notable reductions in GFLOPs and can be combined with other efficient encoder designs to further enhance overall efficiency. {$^\dagger$Task-specific architectures}}
\end{table}
    
\begin{table}[t]
    \begin{center}
        \resizebox{\columnwidth}{!}{
        \begin{tabular}{lcccccccc}
        \toprule
         & \multicolumn{3}{c}{\textbf{Performance} ($\uparrow$)} && \multicolumn{2}{c}{\textbf{GFLOPs} ($\downarrow$)} \\ \cline{2-4} \cline{6-7}
        \multirow{-2}{*}{\textbf{Model}} & \textbf{PQ} & \textbf{mIoU}$_p$ & \textbf{AP}$_p$ && \textbf{Total} & \textbf{Encoder} \\ 
        \midrule
        
        \multicolumn{7}{c}{\textbf{Backbone}: SWIN-T}\\
        \hline
        RT-M2F \cite{lv2023detrs} & 59.73 & 77.89 & 31.35 && 	361.10 & 130.00 \\
        Lite-M2F \cite{li2023lite}  & 62.29 & 79.43 & 36.57 && 428.71 & 172.00 \\
        M2F \cite{cheng2021mask2former} & 64.00 & 80.77 & 39.26 && 537.85 & 281.13 \\
        {\ours}($\GPRatio=0.003$) & 64.18 & 80.49 & 39.64 && 507.51  & 250.80 \\
        {\ours}($\GPRatio=0.01$) & 62.09 & 79.58 & 36.04 && 439.67  & 182.95 \\
        Lite-\ours{}  & 62.64 & 79.99 & 36.52 && 412.88 & 156.17 \\
        \hline
        \multicolumn{7}{c}{\textbf{Backbone}: Res50}\\
        \hline
        M2F \cite{cheng2021mask2former} & 61.86	& 76.94	& 37.35 && 524.11 & 281.13  \\
        PEM \cite{cavagnero2024pem} & 61.07 & 77.62 & 34.11 && 236.60 & -- \\
        YOSO$^\dagger$ \cite{hu2023you} & 59.70 & 76.05 & -- && 265.10 & -- \\
        ReMax$^\dagger$\cite{sun2023remax} & 65.40 & -- & -- && 294.70 & -- \\
        {\ours} & 62.20 & 77.34 & 37.21 && 453.50 & 220.59 \\
        \bottomrule
        \end{tabular}}
    \end{center}
    \label{tab:result_cityscapes}
    \caption{\textbf{Evaluation on Cityscapes dataset}. Our method balances strong performance with lower GFLOPs, making it more efficient than other models that excel in only one aspect. {$^\dagger$Task-specific architectures}}
\end{table}

\begin{table}[t]
    \begin{center}
        \resizebox{.9\columnwidth}{!}{
        \begin{tabular}{lccccc}
            \toprule
            Method& $\beta$ & PQ & mIoU$_p$ & AP$_p$ 
     & GFLOPs\\ \hline
         M2F & - & 39.71 & 46.09 & 26.49 & 139.81 \\
         {\ours} & 0.0005 & 38.79 & 44.71 & 24.81 & 127.21 \\ 
         {\ours} & 0.0010 & 38.80 & 44.22 & 24.78 & 118.23 \\ 
         {\ours} & 0.0020 & 38.63 & 44.11 & 24.59 & 116.95 \\
            \bottomrule
        \end{tabular}}
        \end{center}
    \label{tab:ade20k}
    \caption{\textbf{Evaluation on ADE20K dataset}. The performance of {\ours} on the ADE20K dataset highlights its capability to adaptively balance efficiency and effectiveness. (backbone: Res50)}
\end{table}

\noindent As shown in \cref{fig:teaser_combined}~(b), {\ours} accuracy-efficiency trade-off can be better than the baseline models. Detailed comparisons are provided in \cref{tab:result_coco} and \cref{tab:result_cityscapes}, where we benchmark {\ours} against baseline prior works on the validation set of COCO and Cityscapes datasets, respectively. In \cref{tab:result_coco}, we observe that {\ours} effectively reduces computational costs while upholding performance levels in comparison to M2F \cite{cheng2021mask2former} using both SWIN-T \cite{liu2021swin} and Res50 \cite{he2016deep} backbones. Additionally, {\ours} can be seamlessly integrated into efficient encoder designs, such as Lite-M2F \cite{cheng2021mask2former} \cite{li2023lite}, further reducing GLOPs by approximately 12.6\%. With Res50 as the backbone, MF \cite{cheng2021per}, YOSO \cite{hu2023you}, and RAP-SAM \cite{xu2024rap} underperform compared to {\ours}. While ReMax \cite{sun2023remax} shows competitive accuracy, its specialization in panoptic segmentation limits its general applicability. Our work, however, aims for a broader impact by creating efficient segmentation architectures that can be used for various segmentation tasks and enhance the parent architecture without the size constraints of ReMax.
We observe similar results on the Cityscapes dataset (see \cref{tab:result_cityscapes}) and on the ADE20K dataset \cite{8100027} (as shown in \cref{tab:ade20k}). 

\subsection{Ablation Studies}

% \begin{itemize}[topsep=0.0em,leftmargin=*]
% $\bullet$ \textit

\paragraph{Impact of Adaptation Factor $\GPRatio$} We analyze the impact of $\GPRatio$ on \ours{} and present our analysis in \cref{tab:result_beta}. As expected, a smaller $\GPRatio$ prioritizes segmentation quality over computations, resulting in superior performance. Conversely, a larger $\GPRatio$ signifies a greater emphasis on GFLOPs. This results in a slight sacrifice in PQ leading to a significant reduction in GFLOPs.

\begin{table}[t]
    \begin{center}
        \resizebox{.98\columnwidth}{!}{
        \begin{tabular}{cccccccccc}
            \toprule
             && \multicolumn{3}{c}{\textbf{Performance} ($\uparrow$)} && \multicolumn{2}{c}{\textbf{GFLOPs} ($\downarrow$)} \\ \cline{3-5} \cline{7-8}
            \multirow{-2}{*}{\textbf{Data}} & \multirow{-2}{*}{$\beta$} & \textbf{PQ} & \textbf{mIoU}$_p$ & \textbf{AP}$_p$ && \textbf{Total} & \textbf{Encoder} \\ 
            \midrule
            % \rowcolor{gray!20}
            % \cellcolor{white}
            \multirow{5}{*}{\rot{COCO}} & Baseline & 52.03 & 62.49 & 42.18 && 235.57  & 121.69 \\
            %
            % Exp 00254 ($\alpha=0$) | GFlops = 220.6122 | TxEncGFlops = 107.1847 | PQ = 52.2429 | AP = 41.6080 | mIoU = 62.9488
            % & ($\GPRatio=0$) & 52.24 & 62.95 & 41.61 && 220.61(-6.35) & 107.18(-11.92) \\
            & 0.0 & 52.24 & 62.95 & 41.61 && 220.61 & 107.18 \\
            % Exp 00302 ($\alpha=0.0005$) | GFlops = 202.3905 | TxEncGFlops = 88.4652 | PQ = 52.0643 | AP = 41.5081 | mIoU = 62.7583
            % & ($\GPRatio=0.0005$) & 52.06 & 62.76 & 41.51 && 202.39(-14.08) & 88.47(-27.30) \\
            & 0.0005 & 52.06 & 62.76 & 41.51 && 202.39 & 88.47 \\
            % & ($\GPRatio=0.001$) & 51.72 & 62.60 & 41.12 && 193.10(-18.03) & 79.18(-34.93) \\
            & 0.001 & 51.72 & 62.60 & 41.12 && 193.10 & 79.18 \\
            % Exp 00294 ($\alpha=0.002$) | GFlops = 189.3750 | TxEncGFlops = 75.4496 | PQ = 51.5881 | AP = 40.9104 | mIoU = 62.4907
            % Exp 00301 ($\alpha=0.004$) | GFlops = 185.2981 | TxEncGFlops = 71.3728 | PQ = 51.3246 | AP = 40.5762 | mIoU = 62.4847
            % Exp 00297 ($\alpha=0.02$) | GFlops = 181.6398 | TxEncGFlops = 67.7144 | PQ = 50.7920 | AP = 39.7134 | mIoU = 62.2507
            % & ($\GPRatio=0.02$) & 50.79 & 62.25 & 39.71 && 181.64(-22.89) & 67.71(-44.36) \\
            & 0.02 & 50.79 & 62.25 & 39.71 && 181.64 & 67.71 \\
            \midrule
            % \rowcolor{gray!20}
            % \cellcolor{white}
            \multirow{6}{*}{\rot{Cityscapes}} & Baseline & 64.00 & 80.77 & 39.26 && 537.85 & 281.13 \\
            % Exp 00281 (a=0) | GFlops = 536.0855 | TxEncGFlops = 279.3679 | PQ = 64.5829 | AP = 40.3054 | mIoU = 80.3542 @ 7 / 47 K itr
            % & ($\GPRatio=0$) & 64.58 & 80.35 & 40.31 && 536.09(-0.33)  & 279.37 \\
            & 0.0 & 64.58 & 80.35 & 40.31 && 536.09 & 279.37 \\
            % Exp 00286 ($\alpha=0.001$) | GFlops = 524.0023 | TxEncGFlops = 267.2847 | PQ = 64.4882 | AP = 40.3634 | mIoU = 80.2222
            % Exp 00287 ($\alpha=0.002$) | GFlops = 523.4989 | TxEncGFlops = 266.7812 | PQ = 64.4573 | AP = 39.9118 | mIoU = 79.3287
            % Exp 00308 ($\alpha=0.003$) | GFlops = 507.5138 | TxEncGFlops = 250.7961 | PQ = 64.1835 | AP = 39.6417 | mIoU = 80.4942
            & 0.003 & 64.18 & 80.49 & 39.64 && 507.51  & 250.80 \\
            % Exp 00288 ($\alpha=0.005$) | GFlops = 469.3761 | TxEncGFlops = 212.6585 | PQ = 63.2361 | AP = 38.3008 | mIoU = 79.7302
            & 0.005 & 63.24 & 79.73 & 37.97 && 469.38  & 212.66 \\
            % Exp 00284 ($\alpha=0.01$) | GFlops = 439.6715 | TxEncGFlops = 182.9539 | PQ = 62.0906 | AP = 36.0428 | mIoU = 79.5810
            & 0.01 & 62.09 & 79.58 & 36.04 && 439.67  & 182.95 \\
            % Exp 00285 ($\alpha=0.1$) | GFlops = 411.9808 | TxEncGFlops = 155.2632 | PQ = 60.7127 | AP = 33.8553 | mIoU = 78.1504
            & 0.1 & 60.71 & 78.15 & 33.86 && 411.98  & 155.26 \\
            \bottomrule\vspace{0em}
        \end{tabular}} 
    \end{center}
    
    \caption{\textbf{Impact of the adaptation factor $\GPRatio$}. As the value of $\GPRatio$ increases, the model places greater emphasis on reducing GFLOPs over performance both in COCO and Cityscapes datasets. Baseline here is M2F \cite{cheng2021mask2former}. (Backbone: SWIN-T)}
    \label{tab:result_beta}
\end{table}

\begin{figure}[t]
    \centering
    \includegraphics[width=0.75\columnwidth]{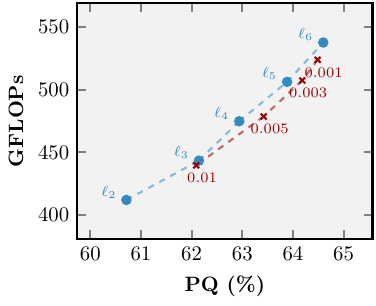} 
    \caption{{\ours} outperforms M2F with WSD by effectively reducing computation without compromising performance, as shown by the varying results with the Gating module across different $\beta$ values. (Dataset: Cityscapes)}
    \label{fig:WSD_layer}
\end{figure}

\paragraph{{\ours} {\vs} M2F w/ Weighted Stochastic Depth (WSD)} We present a comparison between M2F trained with WSD and {\ours} in \cref{fig:WSD_layer}. The blue dots denote the results when M2F w/ WSD exits always at $i$th layer ($i=2, \cdots, 6$). The red ``x''s indicate the performance of our approach with the introduction of the Gating module, which varies with changes in $\beta$. The results clearly demonstrate that {\ours} effectively self-selects encoder layers for compute reduction while preserving performance, illustrating its efficiency in optimizing the trade-off between computation and accuracy.

\paragraph{Impact of Target and Loss Settings for Gating Network Training} We investigate various target and loss settings during the training of the gating network. Specifically, we compare the approach detailed in \cref{sec:training}, using one-hot target and cross-entropy loss (referred to as ``hard-CE'' in \cref{tab:tgt_loss}), with three alternative methods that do not involve setting a specific target exit for each image.

First, we consider using cross-entropy loss between the output of the utility function $\uF(\cdot)$ and the predicted logit passed through a softmax function (referred to as ``u-CE''), \ie, 
\[
\loss_\text{gating}=\sum_i^N\sum_{\Glayer}^{\NGlayer}\uF\superIdx(\Glayer)\ln [\text{softmax}(g_\Glayer\superIdx)]\,.
\]

Second, we apply a softmax function to the utility function $\uF(\Glayer)$ and use cross-entropy as the loss function (referred to as ``soft-CE''), \ie,
\[
\loss_\text{gating}=\sum_i^N\sum_{\Glayer}^{\NGlayer}\text{softmax}(\uF\superIdx(\Glayer))\ln [\text{softmax}(g_\Glayer\superIdx)]\,.
\]

Third, we apply a softmax function to the utility function, but use mean squared error (MSE) loss instead (referred to as ``soft-MSE''), \ie,
\[
\loss_\text{gating}=\sum_i^N\sum_{\Glayer}^{\NGlayer}\left[\text{softmax}(\uF\superIdx(\Glayer))-\text{softmax}(g_\Glayer\superIdx)\right]^2\,.
\]

The analysis in \cref{tab:tgt_loss} is conducted using the SWIN-T \cite{liu2021swin} backbone on the COCO dataset. We observe that ``hard-CE'' yields the most favorable results. As a result, all of the other results use this approach.

\begin{table}
\begin{center}
    \resizebox{\columnwidth}{!}{
    \begin{tabular}{lcccccccc}
    \toprule
     & \multicolumn{3}{c}{\textbf{Performance} ($\uparrow$)} && \multicolumn{2}{c}{\textbf{GFLOPs} ($\downarrow$)} \\ \cline{2-4} \cline{6-7}
    \multirow{-2}{*}{\textbf{Method}} & \textbf{PQ} & \textbf{mIOU}$_p$ & \textbf{AP}$_p$ && \textbf{Total} & \textbf{Encoder} \\ 
    \midrule
    % \rowcolor{eccvblue!20}
    hard-CE & 52.06 & 62.76 & 41.51 && 202.39 & 88.47 \\
    % Exp 00362 52.1621 | mIoU = 62.5833 | AP = 41.5721 | GFlops = 207.4906 | TxEncGFlops = 94.0630 
    u-CE & 52.16 & 62.58 & 41.57 && 207.49 & 94.06 \\
    % Exp 00362V2 (soft V2 tgt $\beta=$5e-4):  | PQ = 51.6382 | mIoU = 62.7542 | AP = 40.8838 | GFlops = 202.0765 | TxEncGFlops = 87.8465
    soft-CE & 51.64 & 62.75 & 40.88 && 202.08 & 87.85 \\
    % Exp 00362V3 (v3):  | PQ = 51.5355 | mIoU = 62.7338 | AP = 40.9113 | GFlops = 198.4558 | TxEncGFlops = 84.5304
    soft-MSE & 51.54 & 62.73 & 40.91 && 198.46 & 84.53 \\
    \bottomrule
    \end{tabular}}
\end{center}

\caption{\textbf{Impact of target and loss in gating network training}. We use ``hard-CE'' loss for training our gating network in all of the other results. (Backbone: SWIN-T; Dataset: COCO)}
\label{tab:tgt_loss}
\end{table}

\begin{figure}[ht]
    \centering
    \includegraphics[width=0.85\columnwidth]{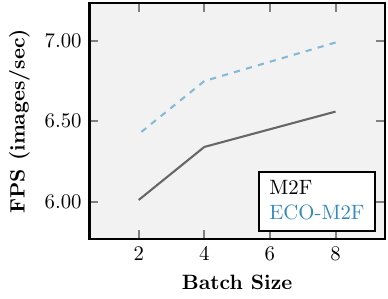} 
    \caption{\textbf{FPS vs. batch size.} {\ours} consistently achieves higher FPS than M2F \cite{cheng2021mask2former} across all batch sizes. (Dataset: Cityscapes; Backbone: Res50)}
    \label{fig:batch}
\end{figure}

\paragraph{Batch Inference and FPS Comparison} {\ours} maintains efficient batch processing by utilizing a shared decoder across all encoder exit points, ensuring that throughput is not compromised. As shown in \cref{fig:batch}, our method significantly improves the frames per second (FPS) across various batch sizes, demonstrating a clear advantage in processing speed. 

\section{Conclusions}
In this paper, we propose an efficient transformer encoder design {\ours} for the Mask2Former-style frameworks. \ours provides a three-step training recipe that can be used to customize the transformer encoder on the fly, given the input image. The first step involves training the parent model to \textit{be dynamic} by allowing stochastic depths at the transformer encoder. The second step involves creating a derived dataset from the training dataset which contains a pair of image and layer number that provides the highest segmentation quality. Finally, the third step involves training a gating network, whose function is to decide the number of layers to be used given the input image. Extensive experiments demonstrate that {\ours} achieves significantly reduced computational complexity compared to established methods while maintaining competitive performance in universal segmentation. Our results highlight {\ours}'s ability to dynamically trade-off between performance and efficiency as per requirements, showcasing its adaptability across diverse architectural configurations, and can be applied to models for object detection tasks.
\paragraph{Limitations.} While {\ours} offers dynamic trade-offs between performance and efficiency according to specific needs, the adaptation factor $\GPRatio$ is a hyperparameter that needs separate tuning for each use case. This is because it relies on the model configuration and dataset characteristics.

{
    \small
    \bibliographystyle{ieeenat_fullname}
    \bibliography{main}
}

\end{document}